%% file: icws.tex
\documentclass[conference]{IEEEtran}
\IEEEoverridecommandlockouts
\usepackage{cite}
\usepackage{amsmath,amssymb,amsfonts}
\usepackage{algorithmic}
\usepackage{graphicx}
\usepackage{textcomp}
\usepackage{xcolor}
\usepackage{booktabs}
\def\BibTeX{{\rm B\kern-.05em{\sc i\kern-.025em b}\kern-.08em
    T\kern-.1667em\lower.7ex\hbox{E}\kern-.125emX}}

\usepackage[ruled,linesnumbered]{algorithm2e}
\usepackage{subfigure}
\usepackage{hyperref}
\usepackage{cleveref}
\newcommand{\nosection}[1]{\vspace{1.5pt}\noindent\textbf{#1.}}
\begin{document}

\title{Federated Learning on Non-iid Data via Local and Global Distillation}

\author{
\IEEEauthorblockN{Xiaolin Zheng$^1$, Senci Ying$^1$, Fei Zheng$^1$, Jianwei Yin$^1$, Longfei Zheng$^2$, Chaochao Chen$^1$\IEEEauthorrefmark{1}\thanks{\IEEEauthorrefmark{1}Corresponding author.}, Fengqin Dong$^3$}
\IEEEauthorblockA{
$^1$\textit{College of Computer Science and Technology, Zhejiang University}, Hangzhou, China\\
$^2$\textit{Ant Group}, Hangzhou, China\\
$^3$\textit{Department of Endocrinology and Metabolism, the First Affiliated Hospital}, Zhejiang University School of Medicine\\
\{xlzheng, scying, zfscgy2, zjuyjw\}@zju.edu.cn, zlf206411@antgroup.com, \{zjuccc, dfq918\}@zju.edu.cn}
}

\maketitle

\begin{abstract}
\input{chapters/abstract}

\end{abstract}


\input{chapters/introduction.tex}
\input{chapters/related_work.tex}

\input{chapters/fed_nd.tex}
\input{chapters/experiments.tex}

\input{chapters/conclusion.tex}
\section{Acknowledgement}
This work was supported in part by the “Pioneer” and “Leading Goose” R\&D Program of Zhejiang (No. 2022C01126), and Leading Expert of “Ten Thousands Talent Program” of Zhejiang Province (No.2021R52001).

\bibliographystyle{elsarticle-num}
\bibliography{refs}
\end{document}

%% file: chapters/abstract.tex
Most existing federated learning algorithms are based on the vanilla FedAvg scheme. 
However, with the increase of data complexity and the number of model parameters, the amount of communication traffic and the number of iteration rounds for training such algorithms increases significantly, especially in non-independently and homogeneously distributed scenarios, where they do not achieve satisfactory performance. 
In this work, we propose FedND: federated learning with noise distillation. 
The main idea is to use knowledge distillation to optimize the model training process. 
In the client, we propose a self-distillation method to train the local model. 
In the server, we generate noisy samples for each client and use them to distill other clients. 
Finally, the global model is obtained by the aggregation of local models.
Experimental results show that the algorithm achieves the best performance and is more communication-efficient than state-of-the-art methods.

%% file: chapters/introduction.tex
\section{Introduction}
Federated learning~(FL)~\cite{mcmahan2017communication,konevcny2016federated,konevcny2016federated2} is a machine learning approach that combines different data sources for model training while ensuring data privacy. 
In recent years, it has attracted much research interest and has been widely adopted in various fields such as 
computer vision~\cite{liu2020fedvision,aggarwal2021fedface,yu2019federated,guo2021multi}, 
natural language processing~\cite{tian2022fedbert,ait2022fedqas,DBLP:conf/emnlp/ZhuWHX20}, 
graph learning~\cite{asfgnn,vfgnn}, 
and 
recommender systems~\cite{liang2021fedrec++,DBLP:conf/kdd/MuhammadWOTSHGL20,minto2021stronger,cui2021exploiting,chen2022differential}.
However, compared with classical centralized machine learning, federated learning is more complex and faced more difficult problems~\cite{kairouz2021advances,khan2021federated,li2020federated,chen2020vertically}.
Currently, there are still many challenges that remain to be addressed.
The main challenges include communication overhead, privacy protection, client statelessness, and the heterogeneous data among clients which greatly limits the further development and application of FL.

In this work, we mainly focus on the heterogeneous data or what is known as data non-independently and identically distributed~(non-iid) problem in FL~\cite{zhu2021federated,zhao2018federated,li2021federated}. 
Most FL algorithms are based on the FedAvg~\cite{mcmahan2017communication}, where the clients train a local model, then upload them to the server and get a global model by averaging the client model's parameters. 
But under the non-iid scenarios, these algorithms require more communication rounds to converge and result in poor performance.

There have been some studies trying to address the non-iid issue, which can be divided into three types. 
(1) \textbf{Weight-constraint methods} 
\cite{li2020fedprox,wang2020tackling,DBLP:conf/iclr/AcarZNMWS21,DBLP:conf/iclr/LiJZKD21,yuan2020federated,DBLP:conf/iclr/ReddiCZGRKKM21,xu2021fedcm,DBLP:journals/corr/abs-1910-06378} use regularization, normalization, or other approaches to constrain the parameters of the client's local models, which makes the local training more stable and reduces the bias of the local models.
(2) \textbf{Client-selection methods} \cite{tang2021fedgp,fraboni2021clustered,briggs2020federated,dinh2020federated,chai2020tifl} optimise the client selection strategy by designing specific metrics. 
(3) \textbf{Knowledge distillation methods}
\cite{li2021model,yao2021local,zhu2021data,jeong2018communication,li2019fedmd,itahara2021distillation,lin2020ensemble,wu2021fedkd,hin2021fedhe} use knowledge distillation to reduce the impact of the non-iid data. 
However, most methods only optimize the client model and just constrain the parameters of the local model in some naive ways, e.g., regularization and distillation.
Therefore, they can only achieve minor improvement in the final model performance.

Observing the challenge in the non-iid scenarios and the limitations of the prior work, in this work, we propose a federated learning algorithm with global and local distillation. 
We design two optimization modules in our FL algorithm: the noise-distillation for the server model and the self-distillation for the client model.
\textbf{Module1:} 
In the server, the received client models are usually biased due to the non-iid data, so the naive aggregation method damages the model performance.
To overcome this, we design a pseudo-sample generation module in the server and use an adaptive method to update the pseudo-samples to make them more similar to real samples. 
Since those samples are generated from noise, we call them \textit{noisy sample}s for the rest of the paper.
We use these samples to update the client models via knowledge distillation and then aggregate the local models to obtain the global model.
\textbf{Module2:} 
In the client, the local model is easy to be over-fitted under the non-iid data, and thus we design a self-distillation module for the client. 
During the training, the input samples will obtain three output probabilities by the dropout layers and the previous model.
Then we use these outputs to distill the current model and make the local model more robust.
The distilled global model is better adapted to the full data distribution than the simple averaging model~(FedAvg Based). And we use self-distillation in the client and noise-distillation in the server, so we name this method \textbf{FedSND}.
Extensive empirical studies show that our proposed approach achieves the best final accuracy and is more communication efficient.

Our main contributions are as follows:
\begin{itemize}
    \item We propose local distillation~(self-distillation) and global distillation~(noise-distillation) to address the non-iid data problem in FL. 
    \item In the client, we propose the dropout layer and self-distillation, which makes the local model more robust. 
    \item In the server, unlike the solutions using public or shared datasets for knowledge distillation, our noisy samples only need to be generated randomly and updated adaptively, which can serve as a distillation model without the reliance on additional data.
    \item We use extensive experiments and analysis on different  datasets and different data distributions to validate the performance of the proposed FedSND and compare it with several state-of-the-art methods.
\end{itemize}

%% file: chapters/related_work.tex
\section{Related Work}
The non-iid data seriously affects the communication efficiency  and makes the performance of the federated model much worse.
Therefore, the current ideas are mainly from the above perspective, including the improvement of the client training process and the selection of client models, specifically, there are three major solutions as follows:

\textbf{Weight-constraint.}
These methods reduce the impact of non-iid data by constraining the model weights. 
\cite{li2020fedprox,DBLP:conf/iclr/AcarZNMWS21} add a regularization term to improve the stability of the local training.
\cite{wang2020tackling, DBLP:conf/iclr/LiJZKD21} use normalization to alleviate the feature shift before averaging local models.
Other studies~\cite{DBLP:conf/iclr/ReddiCZGRKKM21,xu2021fedcm, DBLP:journals/corr/abs-1910-06378, yuan2020federated} aggregate global and local gradient information in previous communication rounds and use the momentum to stabilize the training of the federated model.
However, these methods can only alleviate the bias of local model training, and are less helpful for global models, so the improvement of federated model performance is limited.

\textbf{Client-selection.} 
These methods focus on how to select the best participating clients in each global round. 
The naive FedAvg method selects clients randomly.
\cite{tang2021fedgp, fraboni2021clustered,dinh2020federated} use metrics such as correlation or variance to adaptively select appropriate clients to participate in the server aggregation. 
\cite{briggs2020federated,chai2020tifl} use clustering algorithms to select clients, which improves communication efficiency, and the global model obtained by the server in each round is more responsive to global data. 
However, these approaches are not very helpful for local model training, and the local models obtained from different client data training may vary greatly, and the global model obtained after aggregation is less stable.

\textbf{Knowledge distillation.}
Knowledge distillation is often used to accelerate training and improve model robustness. 
\cite{jeong2018communication, hin2021fedhe} obtain soft labels by averaging the output of different client models and use these labels to distill the client models.
\cite{li2019fedmd,itahara2021distillation,lin2020ensemble} bring public datasets to federated learning, they export the outputs of these datasets by the local models and aggregate them in the server to get the global logits, then distill the local models with global logits.
\cite{li2021model, yao2021local, wu2021fedkd} use a teacher model to guide the student model and only upload student models to the server for federated aggregation, which both optimizes local training and improves communication efficiency. 
Other studies like \cite{zhu2021data, zhang2022feddtg} use the adversarial generation network~(GAN) to obtain the noisy samples, then distill the federated models by these samples.

federated distillation will accelerate the training and get a better model, but most methods need additional teacher models or public datasets to distill the local model.
On the other side, methods using GAN need to upload labels to generate noisy samples.

%% file: chapters/fed_nd.tex
\section{Federated Learning with Noise Distillation}
\subsection{Problem Statement}
We first discuss the definition of the FedAvg, which is the basis of most federated learning algorithms, and then point out the problems that exist on the non-iid data.

\subsubsection{Federated Averaging (FedAvg)} 

we define the loss function of a single sample $(x_i, y_i)$ as $f_i(w) = l(x_i, y_i)$, where $w$ is the parameter of the model, and assume that the number of samples held by a single client is $n$.
The objective function of the model on that client can be represented by 
\begin{equation}
    \min_{w}f(w)=\min_w \frac{1}{n}f_i(w).
\end{equation}
let $n_k=|P_k|$ denote the number of samples held by the client, then the global optimization objective function for federated learning is 
\begin{equation}
    \begin{aligned}
        &\min_w f(w)=\sum_{k=1}^{K}\frac{n_k}{n}F_k(w),\\
        &\text{where } F_k(w)=\frac{1}{n_k}\sum_{i\in P_k}f_i(w).
    \end{aligned}
\end{equation}

As can be seen from the definition of the problem, federated learning consists of two main components: client-side local model training and server-side global model aggregation. 
\textbf{Client-side:} 
Different clients use different datasets to train the model with the same structure, so there will be some difference in the model weights among the trained models, which we can call the model's weights-shift. 
What we need is to mitigate this weights-shift.
\textbf{Server-side:} 
Averaging the uploaded client models directly is a simple and effective method, but we would prefer a better method that results in a global model with better performance.

\subsubsection{The Non-iid Problem}
\begin{figure}[ht]
  \centering
  \includegraphics[width=\linewidth]{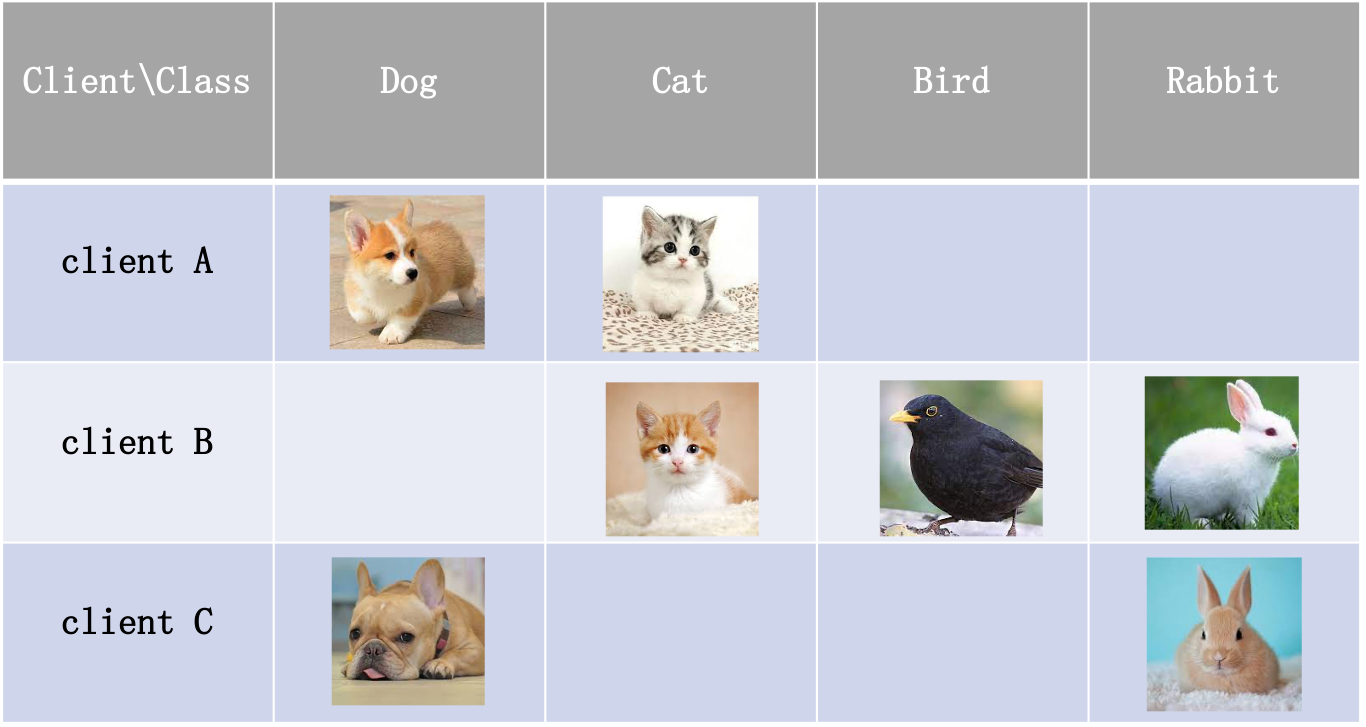}
  \caption{Multiple animal protection organizations need to jointly build machine learning models for animal identification. 
  However, due to environmental and human factors, the animal data collected by different local organizations varies significantly.}
  \label{fig:non-iid}
\end{figure}

When machine learning is performed locally, the data distribution is not a concern, as the train data and the test data usually belong to the same distribution.
However, as shown in \Cref{fig:non-iid}, data distribution among multiple clients can be very different in the federated learning.
Although the federated averaging algorithm is proven to converge in the non-iid scenarios~\cite{DBLP:conf/iclr/LiHYWZ20},
a large number of training iterations is needed. 

Referring to the two parts of FedAvg as discussed above, the impact of non-iid data on federated learning can also be split into two aspects:
\textbf{Client-side:} Compared with the global data distribution, local datasets among clients are very different. And this makes the local model more prone to overfitting.
As shown in \Cref{fig:non-iid}, a single client dataset may only include half of the total classes, thus the trained local model's  generalization capability will be very limited.
\textbf{Server-side:} The weights-shift is even more severe, and the global model obtained by the simple aggregation method is not a good representation of the client models during each communication round.

Therefore, designing suitable methods to solve the above problems can improve the performance of the federated model, and at the same time, improve communication efficiency.

\begin{figure}[ht]
  \centering
  \includegraphics[width=\linewidth]{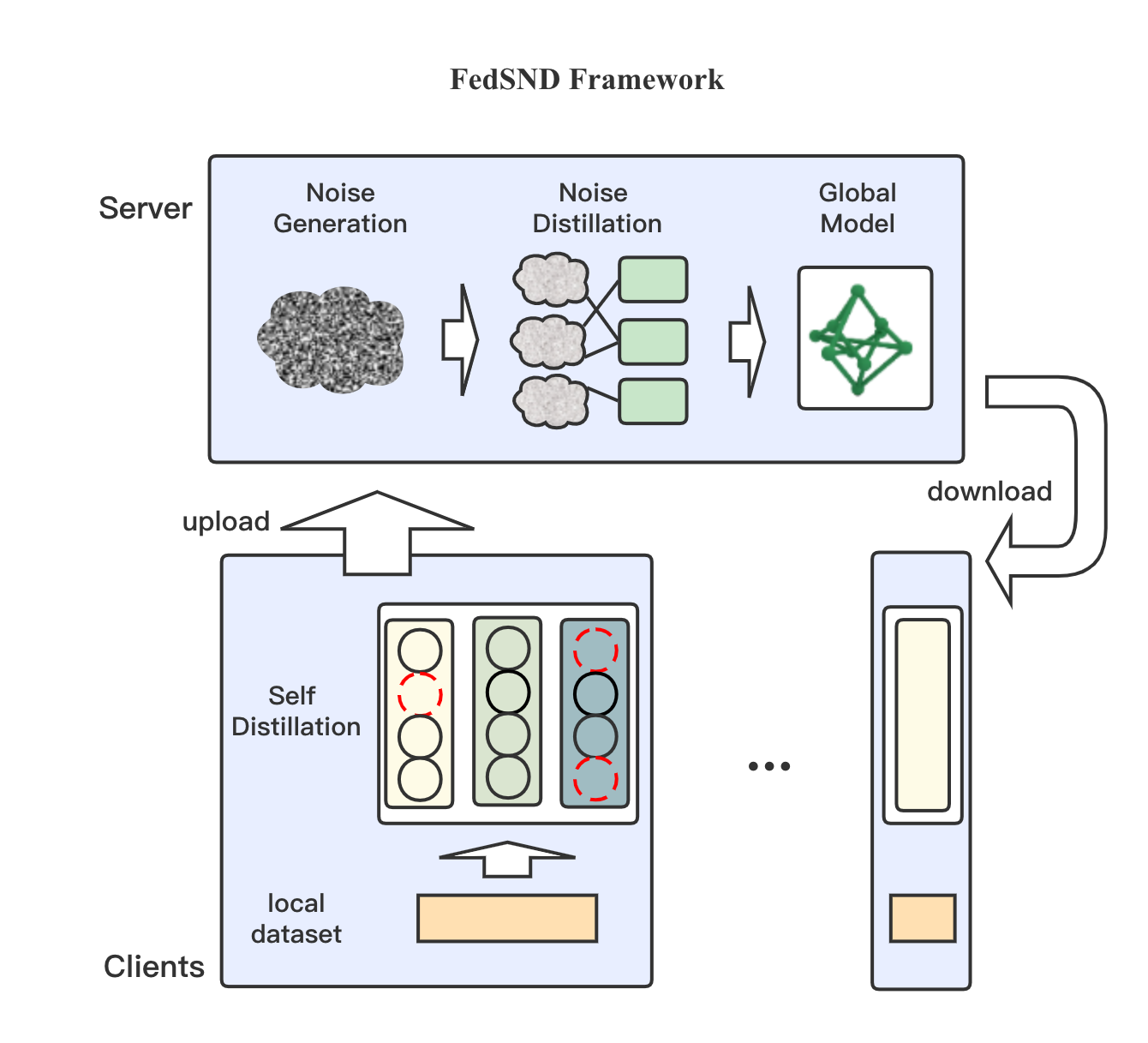}
  \caption{FedSND framework overview: Our algorithm runs in two parts: several clients train the same local model on different datasets, and optimize the training through the proposed self distillation module. After several rounds of training, the client uploads the model to the server. The server distills the received model and aggregates it to generate a global model. Finally, the client downloads the latest global model as a new round of local model to continue training.}
  \label{framework}
\end{figure}
\subsection{Framework Overview}
The FedSND algorithm architecture is based on the vanilla FedAvg scheme, as shown in \autoref{framework}. 

%
%
There are three main modules included.
\begin{itemize}
    \item \textbf{Self-distillation module.} 
    Besides normal training, we use the dropout layer and the last-epoch local model to distill with the local samples. 
    After training a specified number of rounds, the client uploads the latest model's parameters to the server.
          
    \item \textbf{Noisy generation module}, which samples random noise as pseudo-data samples for each client model uploaded to the server.
    However, pure noise is hard to distill.
    Hence, we improve the quality of each client noisy sample by increasing the confidence of the noise samples through an adaptive algorithm.

    \item \textbf{Noise distillation module.}
    After getting the noise samples, we train different client models with other clients' noisy samples to reduce the weights-shift problem for client models via non-iid data.
\end{itemize}

The FedSND framework can well solve the non-iid data problem in federated learning due to the following reasons.
First, the self-distillation module helps the client models overcome overfitting due to data heterogeneity. 
Second, the server's modules generate and use noisy samples to further train the client's weights-shift models with more balanced data, and make the global model become more robust.

\subsection{Self Distillation}
In federated learning, the clients' data is  usually insufficient and unbalanced ~(especially for non-iid data), and the model trained with it is likely to be overfitted, thus affecting the global model performance. 
Knowledge distillation can mitigate overfitting, but distillation methods based on teacher-student models require specific designs for the teacher model. 

We note that if the mechanism of the dropout layer is utilized, distillation can be accomplished using only one model. 
The outputs of the same input sample will be different  when it passes through a model with dropout layers at different times. 
And we can  distill the model by reducing the distance of these outputs, which omits the teacher model in traditional knowledge distillation and alleviates the overfitting problem.

\begin{figure}[ht]
  \centering
  \includegraphics[width=\linewidth]{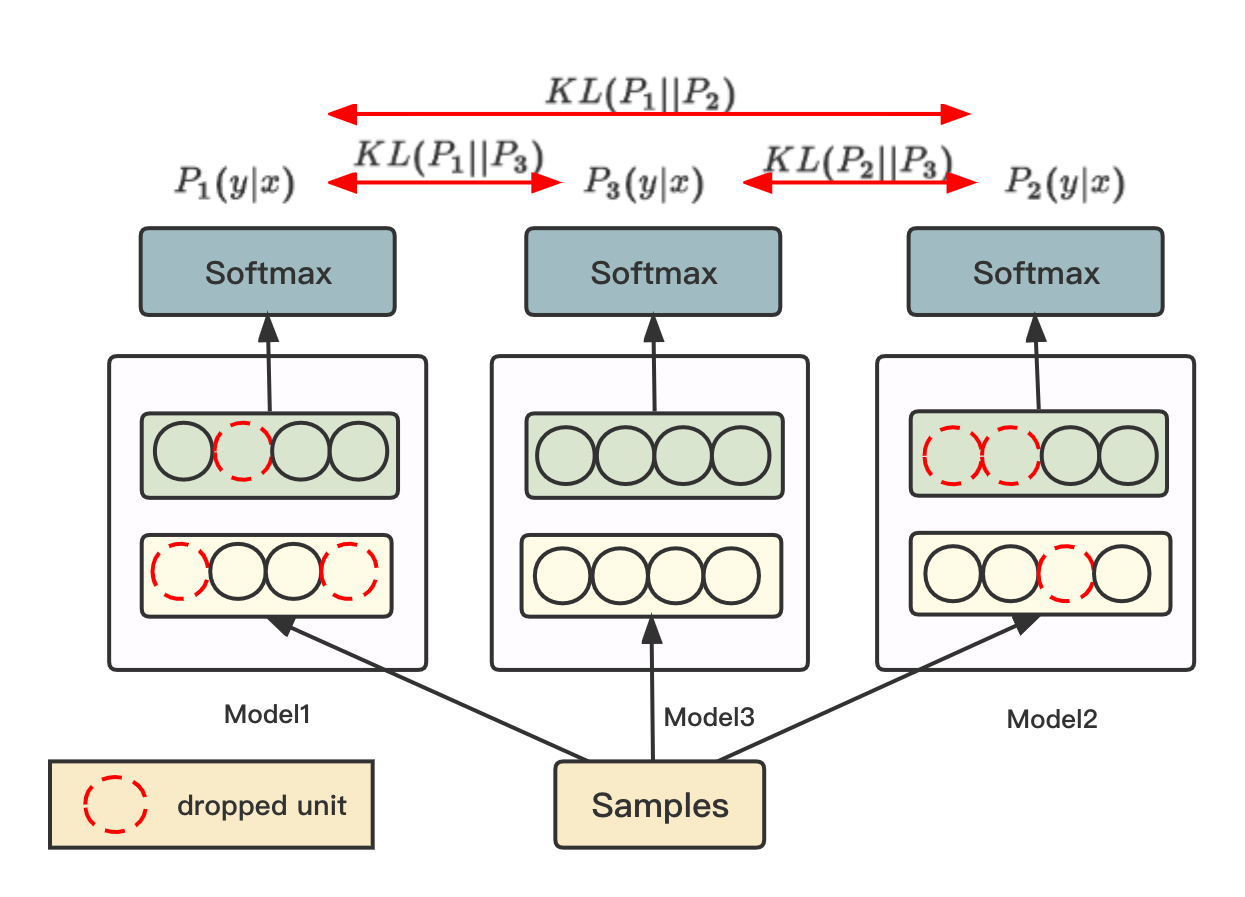}
  \caption{Self Distillation: The client model is composed of three sub-models with the same structure. Two of them are different in the parameters of the dropout layer, while the other model is a model that has been trained in the previous epoch. Samples were simultaneously trained on three sub-models and distilled by KL loss.}
  \label{self-distill}
\end{figure}
Therefore, as shown in \autoref{self-distill}, we have designed a method for local training in the client called self-distillation. 
Specifically, let us take the classification task as an example, the model parameters for the current round are $w_t$, the training samples are $(x, y)$, and the model output is represented by the function $f$. 
We make a copy of the initial model $M_0$ at each round of local training and set it untrainable. 
And the method is constrained by three loss functions. The first is the loss of model outputs to the true labels like:
\begin{equation}
\label{selfloss1}
        L_1 = CE(f_1(w_t;x), y) + CE(f_2(w_t;x), y). 
\end{equation}
Here $CE(\cdot,\cdot)$ means the commonly-used cross-entropy loss.
The probability distributions $f_1(w_t;x)$ and $f_2(w_t;x)$ are obtained by feeding the samples into the model (with dropout) twice. 
Due to the dropout layers in the network, $f_1$ is different from $f_2$. 
The second loss function is defined as follows:
\begin{equation}
\label{selfloss2}
        L_2 = KL(f_1(w_t;x) || f_2(w_t;x)),
\end{equation}
which is to compute the KL divergence between their probability distributions.
Then the samples are passed through the previously fixed model $M_0$ to obtain the probability distribution $f_3$. 
Finally, we compute the distance between the outputs of the current model and the previous model by
\begin{equation}
\label{selfloss3}
        L_3 = KL(f_1(w_t;x) || f_3(w_t;x)) + KL(f_2(w_t;x) || f_3(w_t;x)).
\end{equation}
The total loss is:
\begin{equation}
\label{selfloss4}
        L = \alpha L_1+ \beta L_2+ \gamma L_3, 
\end{equation}
where the $\alpha, \beta, \gamma$ are hyperparameters.

The $L_2$ loss improves the robustness of the local model and reduces the risk of overfitting, while the $L_3$ loss prevents the local model from drifting too far from the previous global model and stabilizes the training process.

\subsection{Noisy sample Generation}
The federated training is difficult due to the distributed data.
An intuitive idea to augment local training is obtaining samples from other clients, but this is not allowed in federated learning scenarios. 

We find that the ultimate goal of data enhancement is to use samples from other clients to make the local model be closer to the others so that when the same sample is passed through different client models, the output probability distribution between them should be as close as possible. 

\begin{figure}[t]
  \centering
  \includegraphics[width=\linewidth]{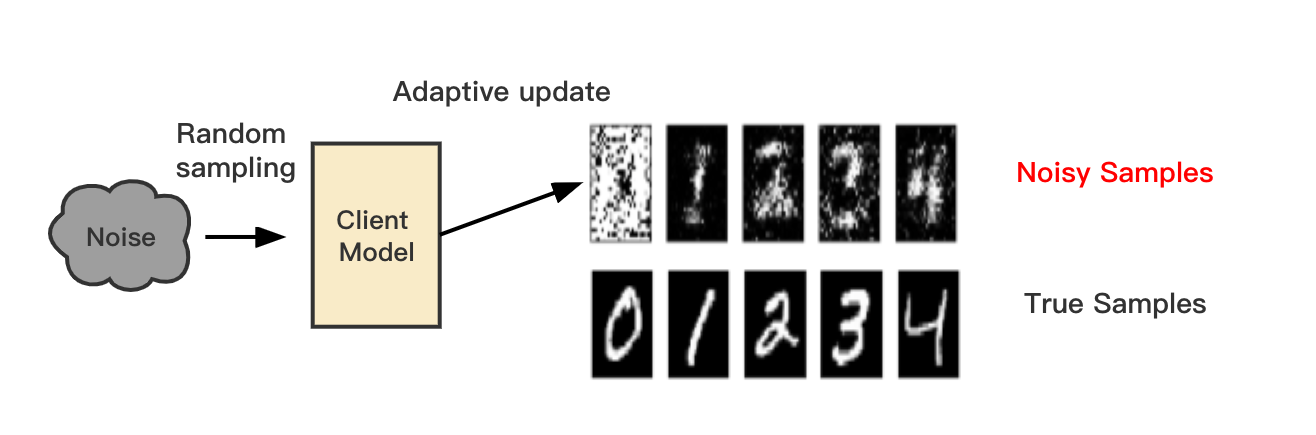}
  \caption{Noise Generation: First, the noise is sampled from the random distribution and output through the client model as a train sample. Then, the constructed loss function $L_e$ is used for reverse updating to obtain a noisy sample that approximates the real sample.}
  \label{noise-gen}
\end{figure}

At this point, the specific value of the sample is no longer important, its main role is to reduce the distance between client models. 
Thus, we propose a method for constructing samples based on random noise which is shown in \autoref{noise-gen}. 

Given any two client models $M_1, M_2$, we use the normal distribution to sample the noisy data as pseudo-samples by 
\begin{equation}
    \begin{aligned}
        &\hat{x_1} \sim \mathcal{N} (\mu_1, \sigma_1), \\
        &\hat{x_2} \sim \mathcal{N} (\mu_2, \sigma_2),
    \end{aligned}
\end{equation}
 and use the pseudo-sample's output probability distribution 
\begin{equation}
    \begin{aligned}
        &\hat{y_1} = f_{M_1}(w_1, \hat{x_1}),\\
        &\hat{y_2} = f_{M_2}(w_2, \hat{x_2}),
    \end{aligned}
\end{equation}
 as the soft label information for the noisy samples.

Although random noisy samples can already distill the model, it is possible that the probability distribution of random noise passing through the model output tends to be uniformly distributed. 
The model does not consider the sample to belong to any  category which makes the KL divergence among different client models is small and distillation may be less effective.

We define $n$ noisy samples with $h$ dim features as $\hat x\in \mathbb{R}^{n\times h}$ and the task is $c$ classification. 
The client model parameters are $w\in \mathbb{R}^{h\times c}$ and let $z$ denote the output of the noisy samples after the model.

To attach probabilistic meaning to the model output, a softmax operation is performed on the output $z$ to obtain the normalized output probability distribution $p$. 
Then we can define the confidence loss function $L_e$ of the noisy samples for the current model by 
\begin{equation}
    \label{adp1}
    \begin{aligned}
        z = f(w,\hat x), \quad  p_i = \frac{e^{z_i}}{\sum_{j}e^{z_j}}, \quad
        L_e=\sum_{i=1}^c p_i\log p_i.
    \end{aligned}
\end{equation}

From the above equation, it can be seen that the smaller the confidence loss, the more effective the model distillation is. 
And we filter out the appropriate noise by setting a suitable threshold. 
However, due to the large noise space, it is difficult to quickly sample all the samples that satisfy the threshold. 
Therefore, we consider setting the features of the noisy samples as trainable parameters, and then update the noisy data by deriving the confidence loss function like
\begin{equation}
    \label{adp2}
    \hat x = \hat x -\eta \frac{\delta L_e}{\delta \hat x}.
\end{equation}
Instead of using the adversarial generation network, this method just increases the noisy sample's confidence, the training often requires only a few iterations to obtain suitable noisy samples.

\subsection{Noise Distillation}
In FedAvg, the server only does the simple aggregation of different client models. 
In contrast, in our method, the server has already generated corresponding noisy pseudo-samples for different clients, so that it can use those noisy samples to further train the client models.

The main problem faced by the clients is weights-shift, which is caused by the fact that the client does not have access to information about the dataset on other clients. 

\begin{figure}[t]
  \centering
  \includegraphics[width=\linewidth]{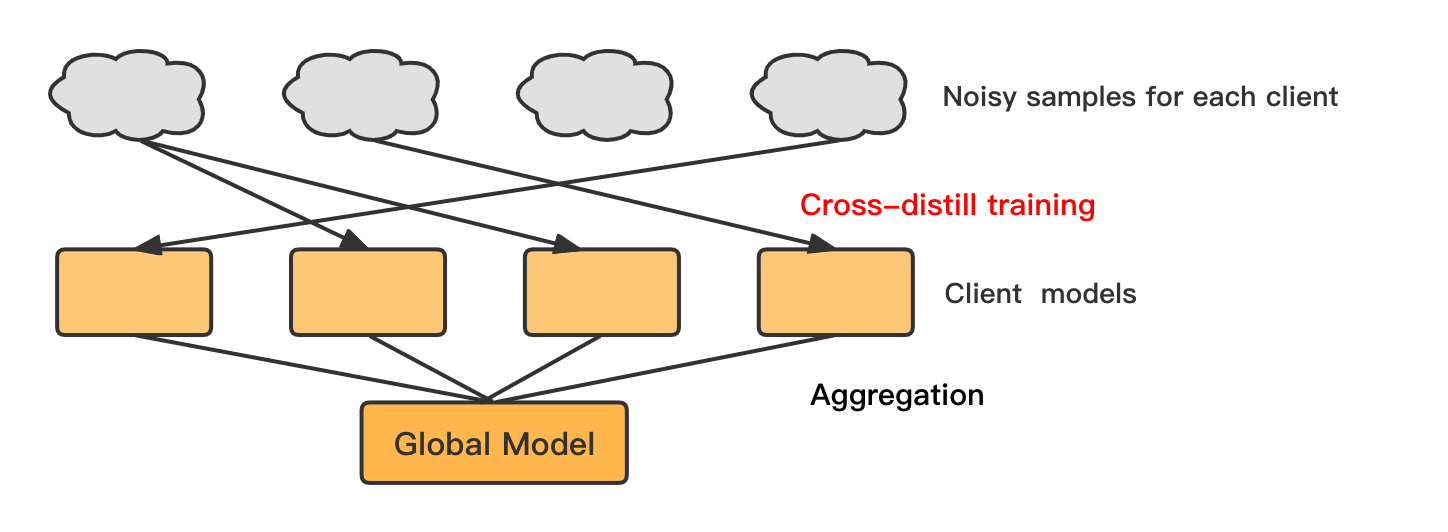}
  \caption{Noise Distillation: The server will first randomly select the client models participating in distillation, and generate noisy samples for each selected model. Then, the noisy samples are used as training data to distill other client models in a cross method. The noisy sample has the information of its own client model and can play the role of models' parameters normalization.}
  \label{noise-distill}
\end{figure}

Our method, as shown in \autoref{noise-distill}, uses noise distillation to overcome this by providing the server with balanced noisy samples to `correct' client models.

Specifically, noise distillation first uses the noisy samples generated by different client models to distill each other models and then averages the distilled client models to obtain the global model. 
\begin{algorithm}[t]
\caption{Noise Distillation in FedSND}\label{algorithm}
\SetKwInOut{Input}{Input}\SetKwInOut{Output}{Output}

\Input{Total $K$ clients, $N$ client participate in noise distillation, model parameter $w$, noisy samples $(\hat{x}, \hat{y})$}
\Output{global model with parameter $w_g$}
\For{$t\leftarrow 1$ \KwTo $K$}{
random sampling $K_c\in \{1, .., K\}$, $|K_c|=N$ \;

\For{$c\leftarrow 1$ \KwTo $K_c$}{
\tcp{get the output for $noise_c$}
$\hat{h_c}= f(w_t, \hat{x}_c)$\; 
\tcp{distill the current $model_t$}
$w_t = \min_{w_t} g(w_t, \hat{h_c}) = KL(\hat{y_c} || \hat{h_c})$\; 
}
\tcp{average the client model parameters}
$w_g = \frac{1}{K}\sum_{t=1}^{K}w_t$
}
\end{algorithm}

We describe FedSND in \autoref{algorithm}. 
Noise distillation allows noisy samples with high confidence generated by a specific client. 
Then the client models will have more similar output distributions to others before aggregation.
The algorithm helps to reduce the difference among the client model parameters and makes the aggregated global model perform better.

\subsection{Summary}
FedSND proposes different optimization strategies for both the client and server sides of federated learning and the whole algorithm is shown in \autoref{algorithm2}. 
we introduce self-distillation on the client to optimize local model training and use noisy samples and noise distillation as described above on the server to obtain a better global model.

\begin{algorithm}[t]
\caption{FedSND: Federated Learning with Self-distillation and Noisy-distillation}
\label{algorithm2}
\SetKwInOut{Input}{Input}\SetKwInOut{Output}{Output}

\Input{Total $K$ clients, the percentage of active clients $C$, client batch size $B$, local epoch $E$, communication round $T$, noise generation threshold $\xi$, model weights $w$}.
\Output{global model $M_g$.}

\SetKwProg{ServerProc}{Server}{:}{}
    \ServerProc{}{
    init model weights $w_0$\;
    \For{$t\leftarrow 1$ \KwTo $T$}{
        $m\leftarrow\max(C*K,1)$\;
        $S_t\leftarrow$ randomly select m clients\;
        \For{\ in parallel\  $k\in S_t$}{
            $w_{t+1}^k\leftarrow$ ClientUpdate($(k,w_t)$)\;
            sampling random noise $\hat{x_k}$, where $L_e(\hat{x_k})<\xi$\;
            adaptive update the noisy samples by equation \eqref{adp1} and \eqref{adp2}\;
        }
        generate the global model $M_g$ by \autoref{algorithm}\;
    }
    }
\SetKwProg{ClientProc}{ClientUpdate}{:}{}
    \ClientProc{$(k_, w)$}{
    divide local dataset by $B$\;
    \For{$e\leftarrow 1$ \KwTo $E$}{
    		copy and fix the previous model $M^{e-1}_0$\;
        \For{$b\leftarrow 1$ \KwTo $B$}{
        		train the local model by equation \eqref{selfloss1}, \eqref{selfloss2}, \eqref{selfloss3}, and \eqref{selfloss4}\;
        }
    }
    }

\end{algorithm}

%% file: chapters/experiments.tex
\begin{figure*}
    \centering

  \includegraphics[width=\linewidth]{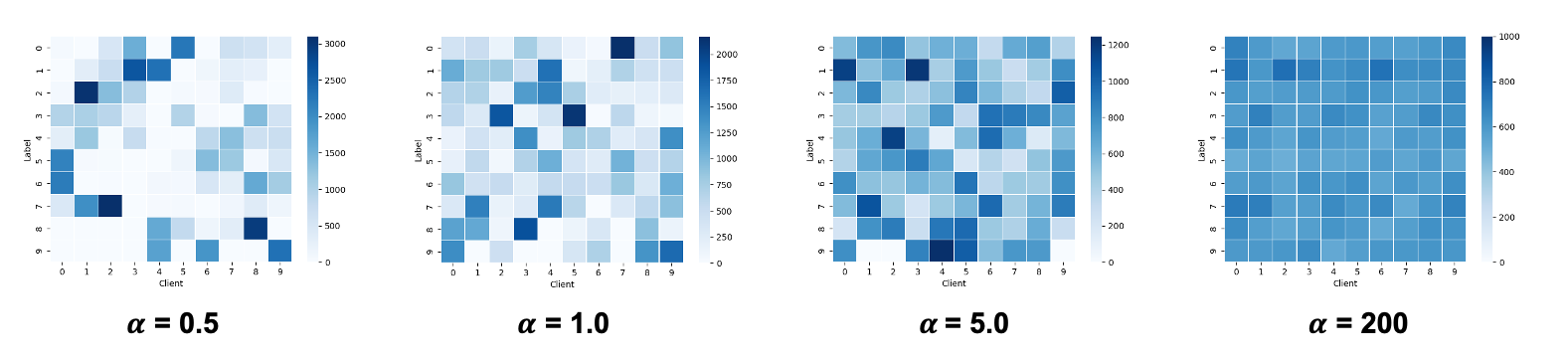}
  \caption{Non-iid Example: Visualization of non-iid data among clients on FashionMNIST dataset, where the x-axis indicates client ids, the y-axis indicates class labels, and the color of the square indicates the number of training samples for a label available to that client.}
  \label{fashionmnist-noniid}

\end{figure*}

\section{Experiments}
We propose to improve the performance and communication efficiency of federated learning through client-side distillation and server-side distillation. 
Specifically, we want to answer the  following questions after our experiments.
\begin{itemize}
    \item[\textbf{Q1:}] How does the FedSND algorithm perform on different datasets?
    \item[\textbf{Q2:}] How efficient is the communication of FedSND on different datasets? 
    \item[\textbf{Q3:}] What is the relationship between the two parts of the FedSND? 
    \item[\textbf{Q4:}] How do some of the hyperparameters of the FedSND in this work affect the experimental results.
\end{itemize}

In order to answer the above questions, we first describe the relevant setup of the experiment.

\subsection{Experimental Setup}
\nosection{Baselines:} 
Our FedSND algorithm uses noisy samples to distill the federated model, so we compare it with vanilla FedAvg and three state-of-the-art approaches based on  knowledge distillation including (1) FedKD\cite{lin2020ensemble}, (2) FedGKD\cite{yao2021local} and (3) FedHe\cite{hin2021fedhe}.

\nosection{Datasets:}
We use four datasets: FashionMNIST, CIFAR-10 datasets, AgNews, and DBPedia for experiments.
The first two datasets are CV (Computer Vision) datasets used for image classification.
FashionMNIST~\cite{xiao2017fashion} is a ten-class gray-scale image dataset that includes images of clothing items such as tops and trousers. 
The dataset contains 60,000 training images and 10,000 test images. 
FashionMNIST is more difficult to train than the classic MNIST dataset while maintaining the same amount of data and format.
CIFAR10~\cite{krizhevsky2009learning} is a colored image dataset. 
Each image is small in size but consists of three channels and is composed of real-world objects including planes, birds, and trucks. 
Due to the large variation in form between objects and the variety of features, a more complex model is required to fit the data. 
The dataset consists of 50,000 training images and 10,000 test images.
The latter two datasets are NLP (Natural Language Processing) datasets used for text classification.
AgNews~\cite{Zhang2015CharacterlevelCN} is a news classification dataset, which is constructed by choosing 4 largest classes from the original corpus. Each class contains 30,000 training samples and 1,900 testing samples.

DBPedia~\cite{2007dbpedia} is a text classification dataset, which is extracted from Wikipedia. 
It contains 342,782 articles and is classified into 9 classes.

\nosection{Non-iid sampling:}
In this work, we need to verify the performance on different data distributions, so we sample existing datasets to generate different distributions for different clients. 
Following the approach proposed by \cite{hsu2019measuring}, we use the Dirichlet distribution function $Dir(p)$ to divide the original dataset.
The non-iid FashionMNIST dataset is visualized in \autoref{fashionmnist-noniid} as an example. 
The x-axis represents the total of 10 clients that participated in the federal training, and the y-axis is the number of labels for each of the corresponding samples. 
It can be seen that at $\alpha = 200$, each client has approximately the same number of labels. 
As $\alpha$ decreases, the difference in the number of labels between the different clients gradually increases. 
At $\alpha = 0.5$, each client essentially has only 2 or 3 major labels, making the training of the federated model more difficult.

\nosection{Configurations:}
we use the classical image classification network Lenet~\cite{lecun1998gradient} as the base client model. 
The extractor is a module with two convolution layers, two  pooling layers, and an activation layer. 
The classifier has two fully connected layers interspersed with activation and dropout layers. 
And all activation functions are ReLU~\cite{DBLP:conf/icml/NairH10}.
In order to facilitate model training and code implementation, some transformations are performed on the input images, including scaling the images to the same size~($28\times 28$), cropping the images randomly to increase the diversity of the images, and using normalization to reduce the cost of model training. 
As for the noisy samples, we randomly initialize the pseudo-samples which have the same shape with raw data and update these samples until the  confidence loss $L_e \leq=e$. 
We set to $\alpha \in \{200,5,1,0.5\}$ to generate non-iid samples among different clients, and the clients are considered to belong to the same distribution when $\alpha = 200$. 
The total number of clients is 100 and setting 20\% active clients at each round. 
We set the client epochs with $epoch_l = 10$, noise threshold $e=0.001$, the number of noisy samples is 50\% of real samples for each client, and make 50\% active clients participate in cross distillation.

\subsection{Accuracy Comparison}
\begin{table*}[]
\centering
\caption{The test accuracy with non-iid parameter $\alpha$ from \{0.5, 1.0, 5.0, 200.0\} on different tasks and datasets.}
\label{accuracy}
\begin{tabular}{ccccccccc}

\toprule
CV                & \multicolumn{4}{c}{FashsionMNIST}                                     & \multicolumn{4}{c}{CIFAR10}                                           \\
methods           & $\alpha=200.0$  & $\alpha=5.0$    & $\alpha=1.0$    & $\alpha=0.5$    & $\alpha=200.0$  & $\alpha=5.0$    & $\alpha=1.0$    & $\alpha=0.5$    \\
\midrule
$\mathrm{FedAvg}$ & 0.8769          & 0.8742          & 0.8713          & 0.8657          & 0.7364          & 0.7217          & 0.7012          & 0.7028          \\
$\mathrm{FedKD}$  & 0.9088          & 0.9075          & 0.8993          & 0.8962          & 0.7889          & 0.7869          & 0.7603          & 0.7424          \\
$\mathrm{FedGKD}$ & 0.8923          & 0.8946          & 0.8951          & 0.8944          & 0.7737          & 0.7822          & 0.7666          & 0.7567          \\
$\mathrm{FedHe}$  & 0.9034          & 0.907           & 0.9024          & 0.8955          & 0.7533          & 0.7587          & 0.7558          & 0.7425          \\
$\mathrm{FedSND}$ & \textbf{0.915}  & \textbf{0.9118} & \textbf{0.9159} & \textbf{0.9152} & \textbf{0.789}  & \textbf{0.7916} & \textbf{0.8}    & \textbf{0.797}  \\
\midrule
NLP               & \multicolumn{4}{c}{AgNews}                                            & \multicolumn{4}{c}{Dbpedia}                                           \\
methods           & $\alpha=200.0$  & $\alpha=5.0$    & $\alpha=1.0$    & $\alpha=0.5$    & $\alpha=200.0$  & $\alpha=5.0$    & $\alpha=1.0$    & $\alpha=0.5$    \\
\midrule
$\mathrm{FedAvg}$ & 0.8758          & 0.8778          & 0.882           & 0.8826          & 0.9272          & 0.9271          & 0.9303          & 0.9346          \\
$\mathrm{FedKD}$  & 0.9074          & 0.9078          & 0.9028          & 0.9028          & 0.9436          & 0.9515          & 0.9454          & 0.9463          \\
$\mathrm{FedGKD}$ & 0.9051          & 0.9046          & 0.9091          & 0.9008          & 0.9461          & 0.9491          & 0.9486          & 0.9527          \\
$\mathrm{FedHe}$  & 0.9071          & 0.9084          & 0.9105          & 0.9036          & 0.9448          & 0.9508          & 0.9452          & 0.9503          \\
$\mathrm{FedSND}$ & \textbf{0.9267} & \textbf{0.9311} & \textbf{0.9326} & \textbf{0.9318} & \textbf{0.9743} & \textbf{0.976}  & \textbf{0.9788} & \textbf{0.9783}\\
\bottomrule
\end{tabular}
\end{table*}

The results of the accuracy experiments are shown in \autoref{accuracy}, and the table includes four datasets, which we will analyze in the following parts.

(1) Overall, the FedSND algorithm performs optimally for each dataset and data distribution. 
The performance of the other algorithms decreases significantly as the non-iid level increases, while the final accuracy of our solution is more stable.

(2) On the CV dataset, FedSND has a significant advantage over the other algorithms. 
In the case of iid, the performance of each algorithm is not significantly different, and our algorithm outperforms the other algorithms by 1\% and 0.1\% on FashionMNIST and CIFAR10 respectively.
In the case of non-iid, the accuracy of the other algorithms dropped severely, while our algorithm was able to achieve a 2\% and 4\% improvement in model's accuracy respectively. 
In addition, the CIFAR10 dataset is more difficult to train than the FashionMNIST dataset.
The performance of FedSND on CIFAR10 is significantly better than the other algorithms, which shows that our algorithm is better adapted to the difficult task.

(3) On the NLP dataset, the FedSND algorithm also showed some improvement, with 2\% and 3\% accuracy improvements on the two datasets respectively.
The results between different datasets and different data distributions do not differ much from the CV dataset, which also shows that the FedSND algorithm can perform well on tasks in different domains.

The above analysis allows us to answer the question \textbf{Q1}: Our proposed FedSND algorithm outperforms other comparative algorithms on all datasets and is highly adaptive as the differences in data distribution increase.

\subsection{Communication Efficiency}
\begin{figure*}
    \centering
  \includegraphics[width=\linewidth]{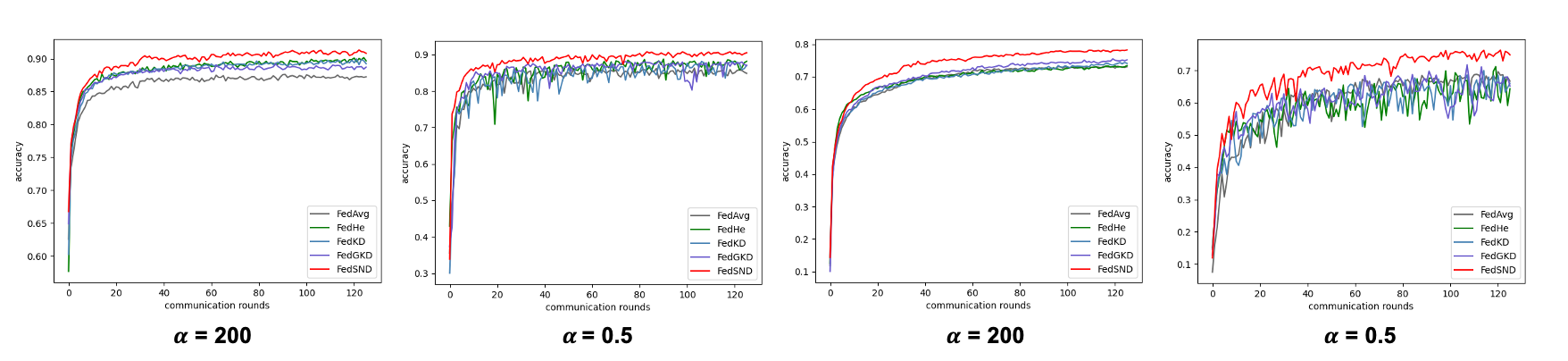}
  \caption{The test accuracy in different number of communication rounds for CV dataset, the two figures on the left are the results of FashionMNIST dataset, and the results of CIFAR10 are on the right.}
  \label{ce-cv}
\end{figure*}

\begin{figure*}
    \centering
  \includegraphics[width=\linewidth]{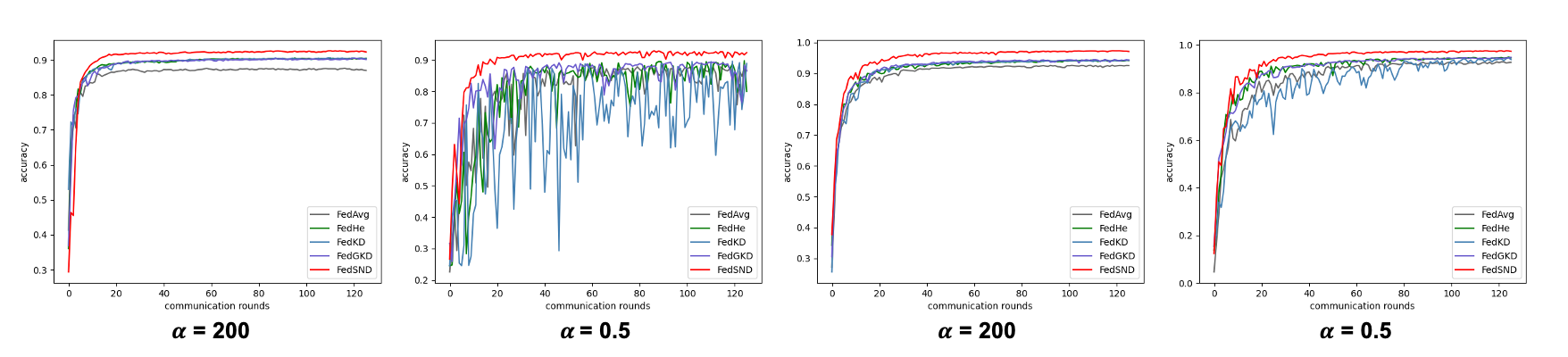}
  \caption{The test accuracy in different number of communication rounds for NLP dataset, the two figures on the left are the results of AgNews dataset, and the results of Dbpedia are on the right.}
  \label{ce-nlp}
\end{figure*}

As convention~\cite{li2021model}, we describe the communication efficiency difference between different models by plotting the number of communication rounds trained by the model and the accuracy rate of each round of the model. 
The higher the communication efficiency of the model, the less the number of communication rounds required to achieve a specific accuracy rate.
We will show the experimental results on CV and NLP datasets in \autoref{ce-cv} and \autoref{ce-nlp} respectively, where the parameter $\alpha = 200$ indicates that the datasets are independent and identically distributed, and $\alpha = 0.5$ indicates that the datasets are non-iid in a very high level.
The following conclusions can be obtained by analyzing the results in the figure:

(1) In general, under the distribution of each data set, the FedSND algorithm has higher communication efficiency than the existing algorithms, and the stability of model training is stronger on the non-independent and identically distributed data sets.

(2) On the FashionMNIST dataset, FedAvg performs worst on the independent and identically distributed datasets, while FedHe and FedKD are more unstable on the non-independent and identically distributed datasets. 
This is mainly because these two algorithms cannot well adapt to the differences in the distribution of different client datasets.
As for CIFAR-10, the curves of each algorithm are more jittery, which indicates that the model is more difficult to learn on this data set.
While the FedSND algorithm is more gentle than other algorithms, which indicates that the algorithm proposed by us is more stable and can adapt to complex training situations.

(3) The performance of the model on the NLP dataset is more unstable. On Agnews, which is not independent and identically distributed, other comparison algorithms are more jittery and have poor stability. On Dbpedia, the model training is stable.
This is because the data volume of the data set is larger, so the model can learn more in the process of one round of communication. 
In the same way, the FedSND algorithm performs better, the model training curve is smoother, and the model can achieve higher accuracy under the same number of communication rounds.
The communication efficiency is higher than other algorithms.

The communication efficiency experiments under different data sets and different data distributions prove that our algorithm FedSND has higher communication efficiency, and with the increase of the difficulty of training data sets, the performance of other algorithms decreases significantly, while our algorithm can ensure certain stability.
Through the above analysis, we can answer the second question \textbf{Q2}: the FedSND algorithm achieves higher accuracy and smoother convergence for the same number of communication rounds, which indicates that self distillation and noisy distillation can reduce the change of clients' data and guide the global model to update in the correct direction.

\subsection{Ablation Study}
\begin{figure*}
    \centering
  \includegraphics[width=\linewidth]{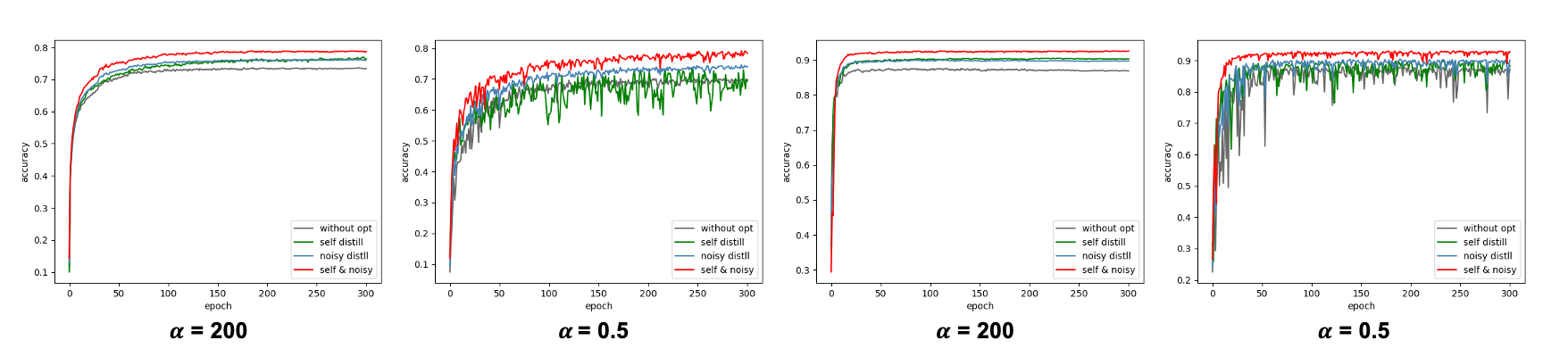}
  \caption{Ablation study on different datasets, the left two figures denote CIFAR10's result and the right two figures denote AgNews's result, we choose two value for $\alpha$, 200(iid) and 0.5 (non-iid).}
  \label{ab-study}
\end{figure*}
\begin{figure*}
    \centering
  \includegraphics[width=\linewidth]{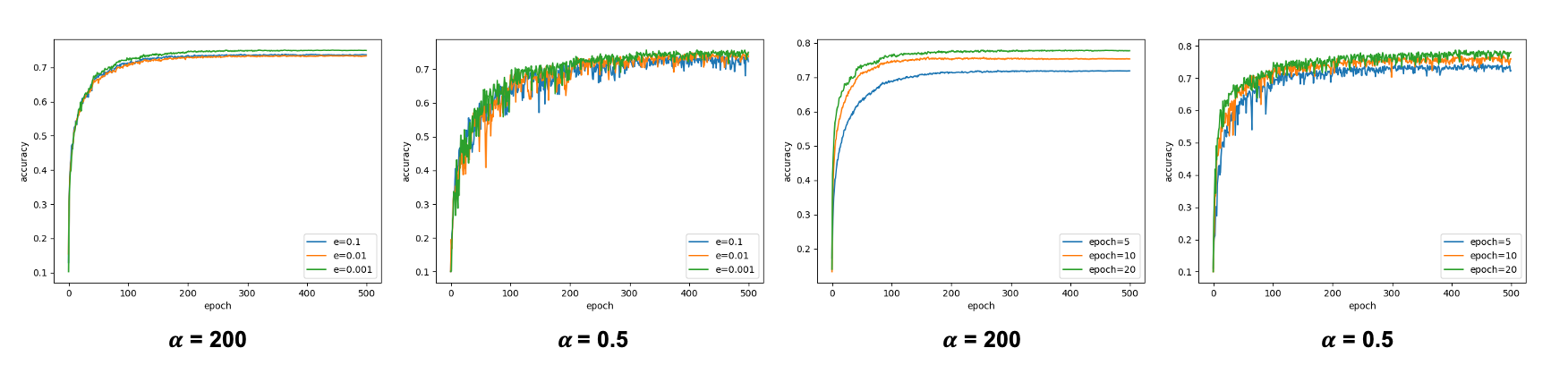}
  \caption{Hyperparameters study on CIFAR10 datasets, the left two figures denote the result of the noise threshold $e$ and the right two figures denote the result of local epochs $epoch$.}
  \label{hyper-study}
\end{figure*}

In our work, FedSND is optimized for both client and server-side, and ablation experiments are needed to verify the contribution of each of the two components, where the client training is optimized by self-distillation and the server is using noisy distillation. 
We set four experiments including (1) without optimization, (2) only self-distillation, (3) only noisy distillation, (4) both two distillation. 
According to the above analysis, the performance of FedSND on different datasets is similar. Therefore, we only selected CIFAR10 and AgNews as representatives for the experiment
and the results are shown in \autoref{ab-study}.

The result shows that the effect of the different modules varies significantly, but in general, they have a beneficial effect on the model training. 
The self-distillation module enhances client training and is more effective in the iid scenario. 
However, in the non-iid scenario, the self-distillation tends to over-fit the local model, which affects the global model and results in exp2 becoming a more oscillating curve in the image.

The noise distillation module is a good way to balance the differences among the clients and provides a more stable training process for the federated model. 
It is also clear from the images that our two optimization methods are complementary to each other. 
Self-distillation is used to speed up the training of the local model and noise distillation is used to smooth the global model, thus making the model training more efficient.

Therefore, for question \textbf{Q3}: both the self distillation and noisy distillation modules play a positive role in the learning of the model. The effect of noise distillation is better, especially in the case of non-iid data, noisy distillation can be a good way to stabilize the model training.

\subsection{Hyperparameters Study}

We found that the model's hyperparameters also have a large impact on the result during the experiment. 
In order to further investigate the role of different hyperparameters, this part conducts experiments on several important parameters of FedSND in two scenarios with $\alpha=\{0.5, 200\}$. 
Due to the similarity of the two datasets, we only choose CIFAR10 dataset and the result is shown in \autoref{hyper-study}.

\textbf{Effect of noisy samples}: The threshold for generating noisy samples is an important hyperparameter. 
We choose three different values $e=\{0.1, 0.01, 0.001\}$. 
The smaller the threshold, the noisy samples will be closer to the true samples. 
Choosing a larger threshold will result in a normal distribution of the noisy sample and will have little effect on the distillation of the model.

The results show that the lower threshold results in higher quality noise and it will guide the model distillation better and make the model converge faster.
However, it can be seen from the images that the accuracy curve does not rise particularly sharply after the reduction of the threshold, suggesting that we do not require the threshold to be set extremely low to have the effect of noise distillation ~(we use $e=0.01$ in previous studies) to reduce the cost of noisy sample generation.

\textbf{Effect of the number of the client local epochs.} The number of client local epochs plays  an important role in federated learning, and the more local iterations of the client model, the better the model fitted on the local data. 
However, due to the data non-iid problem, increasing the number of iterations is not necessarily the best choice for the global model. 
The number of local iterations $epoch = \{5, 10, 20\}$ was set to represent the three cases of under-fitting, fitting and over-fitting.

Based on the experimental results, we can see that the performance of the global model improves as the local epochs increases in the iid scenario. This is because the data distribution between clients is similar at this time, and increasing the number of local epochs gives a more fitted global model.
In the non-iid scenario, the results are less consistent. 
Increasing the number of local epochs does not significantly improve the training results. More local epochs tend to lead to greater variation across client models. 
Therefore, we do not need to set this parameter too large when the data distribution varies widely among clients.

In summary, we can answer the question \textbf{Q4}: an appropriate distillation threshold and the number of client training epochs can improve the quality of the model and promote the performance of the FedSnd algorithm.

%% file: chapters/conclusion.tex
\section{Conclusion}
Federated learning has been widely studied in order to achieve data security and privacy protection in artificial intelligence. 
However, the implementation of federated learning algorithms often requires more resources and studies to reduce the cost of federated learning continue to emerge.

Federated averaging is the most widely used algorithm in federated learning, but the efficiency of the algorithm decreases significantly as the model parameters increase and the differences in data distribution among different data sources. 
In this work, we propose FedSND algorithm based on self-distillation and noise distillation method, which improves the accuracy and the communication efficiency. 
We verify the effectiveness of this algorithm on different data sets and prove the noise distillation is more useful than self-distillation.
In short, we optimize FedSND from both client-side and server-side, and the comprehensive experiments verify its effectiveness.